\documentclass[letterpaper,twocolumn,10pt]{llncs}
\usepackage{multicol}
\usepackage{booktabs}
\usepackage{multirow}
\usepackage{adjustbox}
\usepackage{enumitem}
\usepackage{amsmath}
\usepackage{amssymb}
\usepackage{textcomp}
\usepackage{caption}
\usepackage{array}
\usepackage{siunitx}
\usepackage{subcaption}
\usepackage{graphicx}
\usepackage{threeparttable}
\usepackage{xcolor}
\usepackage{url}
\usepackage{algorithm}
\usepackage{algpseudocode}
\usepackage{orcidlink}

\usepackage[margin=1in]{geometry}
\title{DrugMCTS: a Drug Repositioning Framework Combining Multi-Agent, RAG and Monte Carlo Tree Search}
\author{Zerui Yang\inst{1,2} \orcidlink{0009-0002-9943-250X} 
\and Yuwei Wan \inst{1}
\and Siyu Yan \inst{3}
\and Yudai Matsuda \inst{1}
\and Tong Xie  \inst{4}
\and Linqi Song \inst{1,2}\thanks{Corresponding author}}

\institute{
City University of Hong Kong, Hong Kong, PR China \and 
City University of Hong Kong Shenzhen Research Institute, Shenzhen, PR China \and 
The Hong Kong University of Science and Technology
(Guangzhou), Guangzhou PR China \and 
GreenDynamics, Sydney, NSW, Australia 
}

\begin{document}

\maketitle
 
\section*{Abstract}
Recent advances in large language models have demonstrated considerable potential in scientific domains such as drug repositioning. However, their effectiveness remains constrained when reasoning extends beyond the knowledge acquired during pretraining. Conventional approaches, such as fine-tuning or retrieval-augmented generation, face limitations in either imposing high computational overhead or failing to fully exploit structured scientific data. To overcome these challenges, we propose DrugMCTS, a novel framework that synergistically integrates RAG, multi-agent collaboration, and Monte Carlo Tree Search for drug repositioning. The framework employs five specialized agents tasked with retrieving and analyzing molecular and protein information, thereby enabling structured and iterative reasoning. Extensive experiments on the DrugBank and KIBA datasets demonstrate that DrugMCTS achieves substantially higher recall and robustness compared to both general-purpose LLMs and deep learning baselines. Our results highlight the importance of structured reasoning, agent-based collaboration, and feedback-driven search mechanisms in advancing LLM applications for drug repositioning.
\par{\small\raggedright\noindent\textbf{Code:} \url{https://github.com/yaoge777/DrugMCTS-tool}\par}
\par\noindent\textbf{Contact:} zeruiyang2-c@my.cityu.edu.hk 

\section{Introduction}
Large language models (LLMs) have demonstrated remarkable capabilities across a wide range of domains, including question answering, logical reasoning, and knowledge-intensive tasks such as mathematics and code generation. These models are increasingly being explored for applications in scientific fields, particularly in drug discovery~\cite{ye2025drugassist}. However, when confronted with problems that lie beyond their pre-training knowledge or inherent reasoning abilities, such as predicting novel drug-target interactions, their performance may fall short of expectations~\cite{zheng2025large}.

As general-purpose large models gain traction in scientific domains, fine-tuning on domain-specific datasets has emerged as a common approach~\cite{zhang2024fine,van2025assessment}. However, this method faces several well-recognized limitations. Fine-tuning is computationally expensive and typically domain-specific, making it inefficient and often impractical for extending to new scientific areas. Furthermore, given the dynamic nature of scientific knowledge, maintaining up-to-date models requires continual retraining, which increases computational overhead and risks catastrophic forgetting~\cite{nguyen2019toward}. These challenges hinder the scalability and long-term utility of fine-tuning in real-world scientific applications.

To address the drawbacks of fine-tuning, retrieval-augmented generation (RAG) has emerged as a promising alternative~\cite{zhang2025rag2mol}. By leveraging external agents to retrieve relevant information from literature and databases, RAG enhances the reasoning capabilities of LLMs without requiring model updates~\cite{che2025csstep}. This makes it particularly suitable for dynamic domains such as drug discovery.

Despite its advantages, RAG faces notable limitations. Scientific data can be broadly categorized into structured scientific data and unstructured general-purpose text~\cite{zheng2024large}. Existing RAG systems often prioritize the latter due to better compatibility with LLMs~\cite{song2025llm}, overlooking the richness of structured data such as molecular structures and protein sequences. For example, drug-target interaction tasks frequently exclude structural information, relying solely on knowledge graphs or text~\cite{lee2025rag}, which compromises both prediction reliability and interpretability.

Some methods incorporate domain-specific models to interpret scientific data~\cite{inoue2025drugagent,liu2024drugagent}, but this reintroduces fine-tuning-related limitations. When inputs deviate from training distributions, performance can drop sharply—for instance, drug-target interaction models may lose over 20\% accuracy on unseen molecule-protein pairs~\cite{yang2025mhmg}.

\begin{figure}[!h]
    \begin{minipage}{\columnwidth}
        \centering
        \includegraphics[width=0.9\columnwidth]{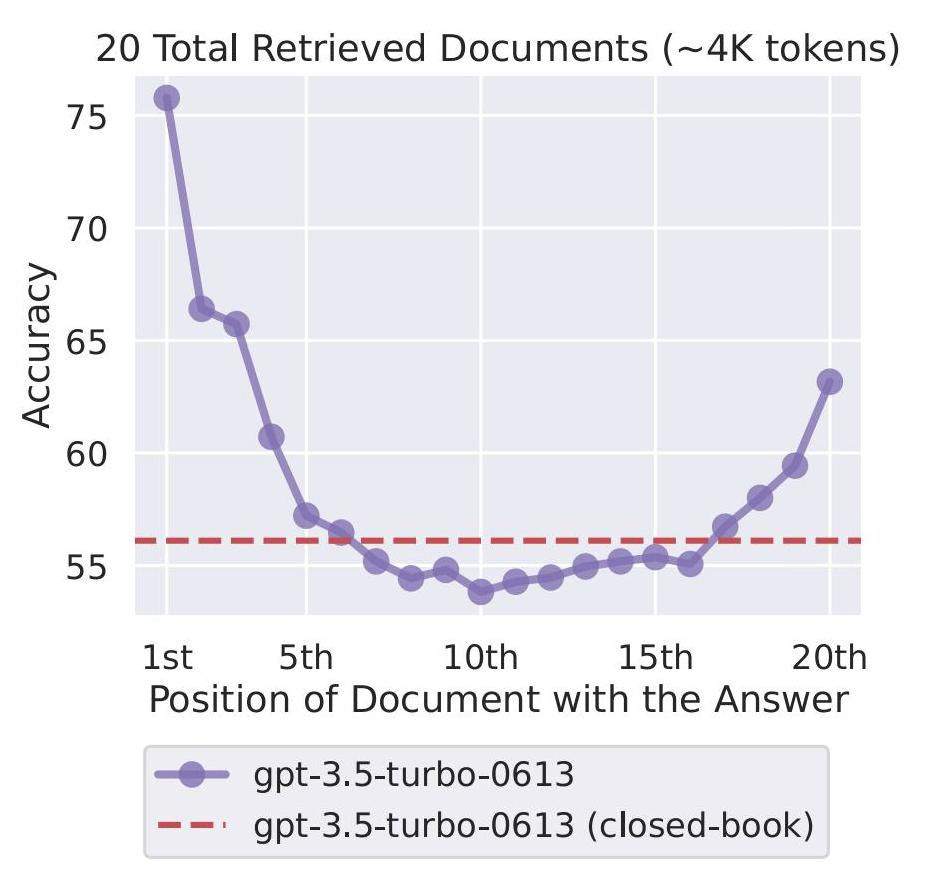}
        \caption{When the document relevant to the query appears in the middle of the retrieved document sequence, the model's performance can even fall below that of the baseline closed-book setting~\cite{liu2023lost}.}
        \label{fig:litm}
    \end{minipage}
\end{figure}

General-purpose data also presents challenges. It often contains noise or errors~\cite{hutter2025lost}, and LLMs may inadvertently discard useful content due to limited contextual understanding. For instance, long-context models are known to suffer from the \textit{“lost in the middle”}~\cite{liu2023lost} phenomenon (Figure~\ref{fig:litm}), where information located in the middle of a long input sequence is more likely to be forgotten or overlooked~\cite{liu2023lost}, leading to incomplete or inaccurate reasoning. Moreover, current approaches typically lack feedback loops, treating drug discovery as a one-shot task rather than an iterative process~\cite{ye2025drugassist,chen2025improving,edwards2024molcap}, which limits robustness and adaptability.

To address the aforementioned limitations, we propose \textbf{DrugMCTS}, a novel drug repositioning algorithm based on RAG, multi-agent collaboration, and Monte Carlo Tree Search (MCTS)~\cite{chaslot2010monte}. DrugMCTS is designed to overcome the challenges posed by computationally expensive fine-tuning, underutilization of structured scientific data, model brittleness in out-of-distribution scenarios, and the lack of iterative feedback mechanisms.

Our system comprises five specialized agents:

\begin{itemize}
    \item \textbf{Retrieval Agent:} Identifies and gathers structurally similar molecules from external scientific databases. While this is the first retrieval step in the pipeline, subsequent agents also incorporate retrieval mechanisms to augment the model’s input with task-relevant scientific knowledge. By leveraging external sources without modifying model parameters, these retrieval-based actions collectively mitigate the limitations of static knowledge and the high cost of fine-tuning.

    \item \textbf{Molecule-Analysis Agent:} Receives a dictionary containing the physicochemical and structural properties of the query molecule and its candidates, obtained from scientific APIs (e.g., RDKit, PubChem). This agent then summarizes the input into a concise natural language description, making the information more accessible to LLMs. 
    
    \item \textbf{Molecule-Selection Agent:} Filters retrieved molecules based on pharmacophore integrity and drug-likeness, effectively eliminating low-relevance or noisy inputs that could compromise model performance. This filtering step also helps mitigate the \textit{“lost in the middle”} phenomenon.

    \item \textbf{Interaction-Analysis Agent:} Analyzes drug-target interaction mechanisms by extracting structured protein data (e.g., pocket features from PDB files via PLIP) and enriching it with natural language summaries from literature. This transformation enables general LLMs to reason over biological structures that they cannot natively parse.
    
    \item \textbf{Decision Agent:} Integrates all upstream outputs—including molecular properties, interaction analyses, and supporting texts—and makes final predictions about likely protein targets. It serves as the final reasoning step informed by the entire multi-agent pipeline.
\end{itemize}

Moreover, the incorporation of MCTS enables the framework to operate with a feedback-driven mechanism, iteratively refining its decision path. This allows the system to explore multiple reasoning trajectories and prioritize more promising nodes based on reward signals, thereby improving both robustness and decision quality over time.

With the guidance of our framework, the lightweight Qwen2.5-7B-Instruct model~\cite{team2024qwen2} achieves performance surpassing Deepseek-R1~\cite{guo2025deepseek}, GPT-4o-mini~\cite{hurst2024gpt} and domain-specific deep learning models on benchmark tasks. Notably, this is accomplished without any domain-specific fine-tuning. We further conduct ablation studies demonstrating that removing any component of our pipeline leads to a performance drop of 1--10\%, underscoring the necessity of each module. 

\paragraph{Software availability.}
The reusable DrugMCTS implementation provides command-line, Python API, and Model Context Protocol (MCP) interfaces for molecular retrieval and MCTS-based target hypothesis ranking. Installation instructions, configuration examples, and implementation differences from the experimental pipeline are documented in the repository. The tool code and synthetic demonstration data are provided under the MIT license; third-party scientific datasets and model weights must be obtained separately under their respective terms. The packaged demonstration uses synthetic data and deterministic model responses. The reusable tool has not been used to revalidate the benchmark results reported in this paper.

\textbf{Our main contributions include:}

\begin{figure*}[!ht]
\centering
\includegraphics[width=\linewidth]{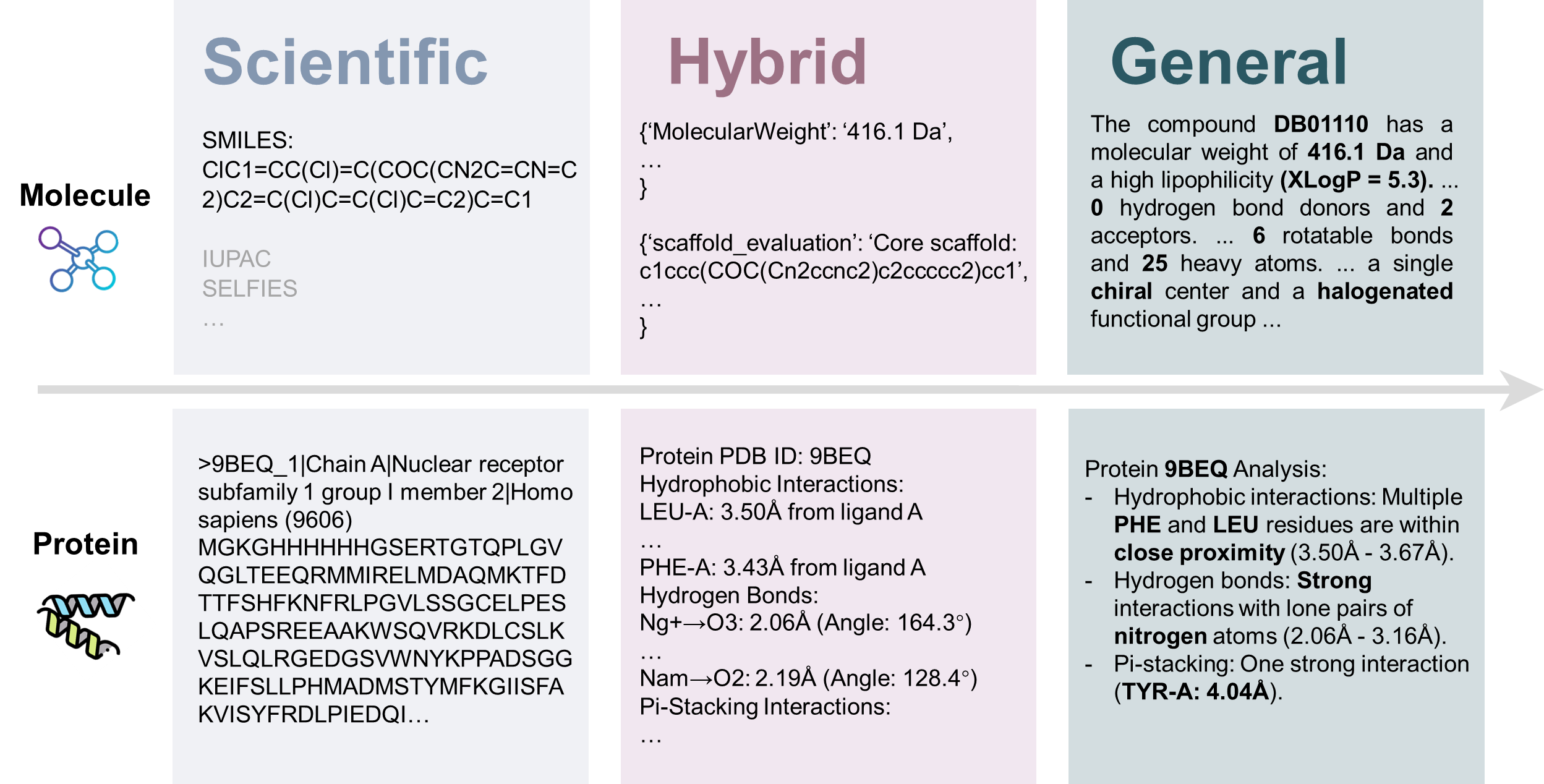}
\caption{Data processing pipiline of DrugMCTS}
\label{fig:datatype}
\end{figure*}

\begin{enumerate}[label=\alph*., leftmargin=*]
    \item We introduce an end-to-end drug repositioning framework that enables Qwen2.5-7B-Instruct to outperform much larger models such as Deepseek-R1, without requiring any fine-tuned domain-specific model. The method leverages external structured and textual knowledge sources to boost reasoning and decision-making capabilities.
    
    \item By incorporating MCTS, our framework introduces a feedback mechanism that enables iterative refinement of decisions, autonomous filtering of noisy inputs, and robust identification of high-value evidence—addressing the lack of adaptability in current one-shot inference systems.
    
    \item We propose a systematic data processing pipeline (Figure~\ref{fig:datatype}) that transitions from structured scientific data to hybrid scientific-general data, and finally to general-purpose language input. This hybrid representation exploits the strengths of each modality, enhances model interpretability, and offers a reusable workflow applicable beyond drug-target interaction tasks.
\end{enumerate}

\section{Method}
\subsection{Overview of DrugMCTS}

\begin{figure*}[!ht]
\centering
\includegraphics[width=\linewidth]{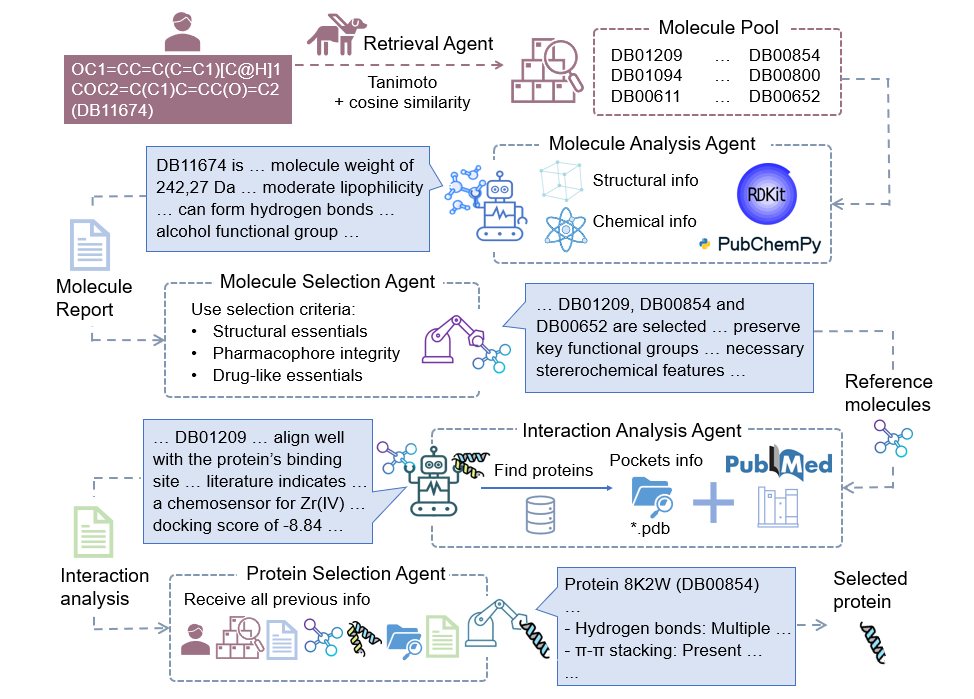}
\caption{Illustration of a single rollout in the DrugMCTS framework. The figure demonstrates how agents sequentially collaborate during one inference trajectory guided by MCTS.}
\label{fig:workflow}
\end{figure*}

Our proposed framework, \textbf{DrugMCTS}, enhances molecular-protein interaction prediction through a structured multi-agent system guided by MCTS, as illustrated in Figure~\ref{fig:workflow} and Appendix Algorithm~\ref{alg:drugmcts}. Given a query molecule $M_{qm}$, DrugMCTS coordinates a sequence of agents to retrieve relevant chemical and biological information, perform structured analysis, and identify promising protein targets.

Each agent contributes domain-specific reasoning: retrieving structurally similar molecules, analyzing molecular properties, filtering irrelevant candidates, and evaluating interaction potential through structural and textual evidence. Instead of relying on end-to-end model predictions, DrugMCTS incrementally builds reasoning paths, ensuring interpretability and robustness at every step.

Crucially, we incorporate MCTS as an inference-time decision mechanism. At each decision point, multiple hypotheses are generated and evaluated using the Upper Confidence Bound applied to Trees (UCT) algorithm (Equation~\ref{eq:uct})~\cite{couetoux2011continuous}, enabling the framework to balance exploration and exploitation. This search-guided process allows DrugMCTS to iteratively refine its predictions and autonomously filter low-quality or noisy information without requiring domain-specific fine-tuning.

By integrating structured data retrieval, expert-agent collaboration, and search-based decision-making, DrugMCTS offers a scalable and interpretable approach for scientific discovery in dynamic domains such as drug-target interaction.

\begin{equation}
\text{UCT} = \frac{W_i}{n_i} + c \sqrt{\frac{\ln N}{n_i}} \label{eq:uct}
\end{equation}

\subsection{MCTS for Drug Repositioning}
\begin{figure}[!h]
    \begin{minipage}{\columnwidth}
        \centering
        \includegraphics[width=0.9\columnwidth]{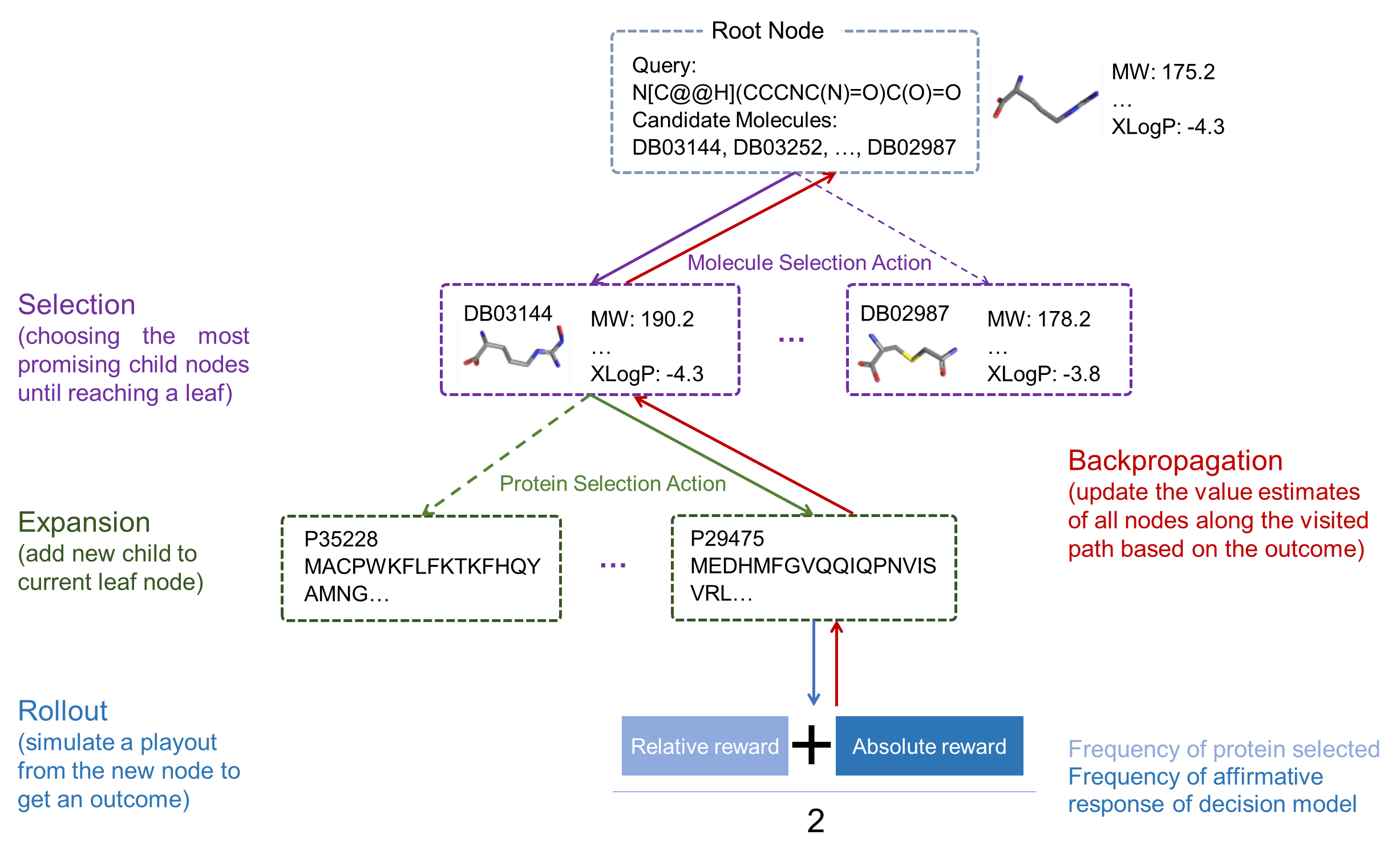}
        \caption{Example of MCTS for drug repositioning}
        \label{fig:mcts}
    \end{minipage}
\end{figure}

Given that MCTS is a well-established algorithm widely adopted and thoroughly explained in various studies~\cite{li2025alpha}, we provide only an overview here. As shown in Figure~\ref{fig:mcts} and Appendix Algorithm~\ref{alg:mcts}, the MCTS algorithm consists of four main steps: selection, expansion, simulation, and backpropagation.

In each rollout, the process begins at the root node and proceeds by selecting a leaf node using the UCT algorithm. The selected leaf node is then expanded by generating one or more child nodes. This selection-expansion cycle continues iteratively until reaching an end node. Upon reaching this terminal node, a predefined scoring rule is applied to evaluate its quality. The score is then backpropagated up the tree to update the scores and visit counts of all nodes along the path. After completing a pre-defined number of rollouts, we obtain a series of candidate solutions from which the final answer is chosen.

\subsection{Action Space}
The action space refers to the set of possible actions that can be performed during each expansion phase. Similar to chain-of-thought (CoT) reasoning, these actions are sequential and interdependent; each subsequent action builds upon the results of the previous one. Therefore, coordination among different agents is essential for coherent execution.

Our framework includes six distinct actions corresponding to five specialized agents. Except for the retrieval agent, which does not invoke any model, all other agents utilize the same LLM without requiring additional fine-tuned models.

\textbf{$A_1$ Retrieval Action.}
Upon receiving the query molecule, the Retrieval agent queries databases to identify molecules structurally similar to the query molecule. We employ two similarity metrics: the Tanimoto coefficient~\cite{bajusz2015tanimoto} and the cosine similarity based on the last hidden state computed by ChemBERTa. For each metric, we retrieve the top-10 most similar molecules, merge the results, and remove duplicates to form the initial candidate molecules $M_{cm}$. The proteins can interact with $M_{cm}$ are also retrieved, forming candidate proteins $P_{cp}$. These molecules serve as inputs to the root node of the MCTS.

Since the similarity scores are deterministic, the resulting molecule pool remains consistent, satisfying the requirement that the root node must be unique.

\begin{equation}
\text{sim}_C(M_{qm}, M_i) = \frac{ \mathbf{h}_{qm} \cdot \mathbf{h}_i }{ \| \mathbf{h}_{qm} \| \cdot \| \mathbf{h}_i \| }
\end{equation}

\begin{multline}
M_{cm} = \text{dedup} \Big( \text{Top}_{10}\left(\text{sim}_T(M_{qm})\right) \\
\cup\ \text{Top}_{10}\left(\text{sim}_C(M_{qm})\right) \Big)
\end{multline}

\textbf{$A_2$ Molecule Analysis Action.}
Understanding molecular properties is crucial to predict molecular-target interactions~\cite{zheng2025large}. While general-purpose LLMs can perform some analysis using SMILES representations, their interpretations are often incomplete and prone to errors, especially when experimental data, such as hydrophobicity, is required.

To address this limitation, the Molecule Analysis (MA) agent first utilizes RDKit and PubChemPy APIs~\cite{swain2014pubchempy} to extract a set of structural and physicochemical properties ($C_{q,s}$ \& $C_{q,phy}$) for the query molecule $M_{qm}$ by calling RDKit and PubChemPy APIs. These properties include:

\begin{itemize}
\item Structural features: chiral centers, scaffolds, and functional groups.

\item Physicochemical properties: molecular weight, lipophilicity (logP), polar surface area (PSA), hydrogen bond donors/acceptors, rotatable bonds, and heavy atom counts.
\end{itemize}

Based on this structured and quantified information, the agent then generates a comprehensive molecular analysis report $R_{qm}$.


\textbf{$A_3$ Molecule Selection Action.}
Research has shown that the quality of retrieved information significantly impacts the accuracy of model-generated answers~\cite{csakar2025maximizing}. Excessive irrelevant information can negatively affect the model’s performance~\cite{luo2025does}. Although the retrieved molecules share structural similarities with the query molecule, they may not necessarily provide useful insights for drug repositioning tasks~\cite{lee2025rag}.
Thus, the Molecule Selection (MS) agent filters the molecule pool, based on structural similarity to the query molecule,  pharmacophore integrity, and drug-like essentials, to generate reference molecules $M_{rm}$. The reference proteins $P_{rp}$ are obtained by selecting the proteins that can interact with $M_{rm}$ from $P_{cp}$.
To ensure that these reference molecules are thoroughly characterized, this action also invokes the \textbf{$A_2$ Molecule Analysis Action}. Specifically, the MA agent is called to retrieve the structural and physicochemical properties ($C_{c,s}$ \& $C_{c,phy}$) of $M_{cm}$ but without generating reports.


\textbf{$A_4$ Interaction Analysis Action.}
In this step, we aim to analyze potential interactions between the $M_{rm}$ and $P_{cp}$. A major challenge lies in interpreting protein structures from amino acid sequences, which general-purpose LLMs struggle to handle due to their complexity.
To overcome this, we adopt the methodology from DrugRealign~\cite{wei2024drugrealign}, utilizing Python’s PLIP library to extract binding pocket information $D_{bp}$ from PDB files and present it in textual format. In contrast to the original tabular representation which can be difficult for LLMs to parse, we reformat each entry into a descriptive paragraph to improve interpretability.
Additionally, we retrieve relevant scientific literature $L_{rp}$ from PubMed~\cite{white2020pubmed} to provide contextual support for interaction analysis by Interaction Analysis (IA) agent.


\textbf{$A_5$ Protein Selection Action.}
At this stage, the Decision agent synthesizes all available information, including the $M_{qm}$, $R_{qm}$, $M_{rm}$, $P_{cp}$, $D_{bp}$, and $R_{ia}$.Based on this integrated knowledge, the agent selects the most promising target protein $P_{s}$ from the full list of candidates.


\textbf{$A_6$ End Action.}
This action does not involve any agent invocation. When the model selects the final protein, the End Action is executed. Upon encountering an end node, the MCTS algorithm terminates further expansion, evaluates the end node’s score, and backpropagates the updated values to all nodes along the traversal path, concluding the current rollout.

\subsection{Reward Calculation}
Predicting molecular-protein binding affinity typically involves two primary approaches: molecular docking methods~\cite{huey2012using} and deep learning-based methods~\cite{yang2025mhmg}.

\textbf{Molecular Docking Methods.}
Molecular docking methods, such as AutoDock Vina~\cite{huey2012using}, are widely used but suffer from notable limitations. First, they require accurate three-dimensional structural information of the target proteins, which is often unavailable or unreliable for many biologically relevant molecules. Second, these methods are computationally intensive, limiting their scalability in large-scale drug repositioning tasks. Given these limitations, there has been a growing interest in developing more efficient alternatives.

\textbf{Deep Learning-Based Methods.}
Deep learning models have emerged as a promising solution due to their superior computational efficiency~\cite{yang2024iresnetdm}. However, their performance is highly dependent on the training dataset. When predicting samples that deviate significantly from the training data, the accuracy of these models can drop by over 20\%~\cite{yang2025mhmg}, limiting their applicability across diverse scenarios.

\subsubsection{Self-Consistency Score.}
To address these challenges, our work adopts an alternative reward calculation method known as the self-consistency score~\cite{wang2022self}. This approach involves querying the model multiple times with the same question and selecting the most frequently occurring answer $p^*$ as the final output. The frequency of this answer serves as the \textbf{relative reward}. 
However, this method introduces a potential issue: if all candidate proteins in one rollout exhibit strong binding affinity with the query molecule, while in another rollout none do, the calculated relative rewards could still be similar despite the stark differences in actual binding affinities. To mitigate this limitation, we introduce an absolute reward mechanism.

\textbf{Absolute Reward.}
The absolute reward is computed by inputting the $P_{s}$, $D_{bp}$, $L_{rp}$, $M_{qm}$, and $R_{qm}$ into a decision-making model. This model evaluates whether there is a significant interaction between the protein and the query molecule. The frequency of affirmative responses ("yes") is then used as the absolute reward.

\textbf{Final Reward Calculation.}
The final reward for each rollout is calculated as the average of the relative reward and the absolute reward. This combined approach ensures that both the consistency and the strength of the predicted interactions are taken into account, providing a more robust evaluation metric.

\begin{equation}
R_{\text{relative}}(p^*) = \frac{\text{Number of times } p^* \text{ is selected}}{\text{Total number of selections}}
\end{equation}

\begin{equation}
R_{\text{absolute}}(p^*) = \frac{\text{Number of "yes" responses}}{k}
\end{equation}

\begin{equation}
R_{\text{final}}(p^*) = \frac{R_{\text{relative}}(p^*) + R_{\text{absolute}}(p^*)}{2}
\end{equation}

\section{Experiments}
\subsection{Datasets and Metrics}
We utilized two datasets, DrugBank~\cite{knox2024drugbank} and KIBA~\cite{tang2014making}, which were processed to include a total of 788 entries from DrugBank and 626 entries from KIBA. Each entry consists of a molecule as input and its corresponding interacting proteins as output. The number of ground truth interactions per entry varies. To evaluate model performance, we used recall, defined as the ratio of correctly predicted proteins to the total number of ground truth proteins. The detailed information of data processing pipeline can be found in Appendix Section~\ref{sec:dataset}

\begin{equation}
\text{Recall} = \frac{|\{\text{proteins predicted correctly}\}|}{|\{\text{all ground truth proteins}\}|}
\end{equation}

\subsection{Settings}
For the retrieval phase, we employed ESM2\_t33\_650M\_UR50D to compute cosine similarity. For all other inference stages, we used Qwen2.5-7B-Instruct. Specifically, each search process involved 12 rounds of rollouts. Except for the protein selection and end actions, which generate only one node during expansion, all other actions generated four nodes with distinct answers per expansion. The temperature was set to 0.8. Both relative and absolute rewards were computed by generating four responses per rollout, also with a temperature of 0.8.

\subsection{Baselines}
We established three sets of baselines to compare our model's performance:

\textbf{General Models. (GM)}
We selected GPT-4o-mini and Deepseek-R1 as general-purpose models. In this group, we provided the models with minimal information: the SMILES representation of the query molecule, reference molecules, and candidate proteins along with their pocket types.

\textbf{General Models with RAG. (GM + RAG)}
Many existing studies on drug repositioning using LLMs either employ divergent methodological formulations or are not open-sourced, which complicates comparative evaluation. Our approach adopts the framework of the DrugReAlign~\cite{wei2024drugrealign} method, with modifications and enhancements tailored to our specific formulation. To ensure equitable comparison, we incorporate not only protein pocket information but also incorporate $C_{q,s}$ and $C_{q,phy}$ as additional features. The models used in this group remained GPT-4o-mini and Deepseek-R1.

\textbf{Deep Learning Models. (DL Models)}
We trained four deep learning models: AttentionDTA~\cite{zhao2019attentiondta}, GraphDTA~\cite{nguyen2021graphdta}, DeepConv\_DTI~\cite{lee2019deepconv}, and Perceiver CPI~\cite{nguyen2023perceiver}, on both DrugBank and KIBA datasets. We extracted data involving the query molecules as test sets and used the remaining data for training. All four models achieved over 70\% accuracy on the test sets. During testing, we applied majority voting to select the final answer by averaging the scores from the four models and choosing the top-k proteins, where k corresponds to the length of the ground truth.

\subsection{Results}
The experimental results (Table~\ref{tab:exp}) indicate that general-purpose LLMs (GPT-4o-mini and Deepseek-R1) achieved relatively low recall scores of only 12.59\%–16.19\% on the DrugBank dataset when operating in a zero-shot setting. However, when incorporating molecular structural features and chemical properties via RAG-based prompting, model performance decreased, with GPT-4o-mini dropping to 15.19\% and Deepseek-R1 significantly falling to 12.59\%. This suggests that the inclusion of potentially irrelevant or misleading information through retrieval can negatively impact the reasoning capabilities of general-purpose LLMs.

Among the deep learning baselines, the ensemble of four specialized models (AttentionDTI, GINConvNet, DeepConv\_DTI, Perceiver CPI) achieved a recall score of 23.64\%, representing an 84\% improvement over the best-performing general-purpose model. On the KIBA dataset, the same ensemble attained a recall score of 26.45\%, further demonstrating its effectiveness in capturing drug-target interactions.

Our proposed method, DrugMCTS, significantly outperformed all baseline approaches. Using a base TopK strategy, DrugMCTS achieved a recall of 44.66\% on DrugBank and 42.24\% on KIBA. This represents improvements of approximately 88.9\% and 31.7\% over the best deep learning baselines, respectively. Furthermore, our dynamic adjustment strategy (TopK+3) boosted performance to 55.34\% on DrugBank and 49.24\% on KIBA, marking maximum improvements of 330\% and 91.4\% over the general-purpose models.  It is worth noting that although the other three methods showed improved performance on the KIBA dataset, our method exhibited a slight drop. However, according to Appendix Table 2, this apparent improvement in other baseline settings is largely due to an increased ratio of ground truth candidates among the total options. As shown in Appendix Table 1, compared to DrugBank, the KIBA dataset contains a larger number of candidate proteins per drug, which increases the difficulty of selecting the correct $P_{cp}$ after molecule selection. This observation indirectly highlights the importance of effective filtering mechanisms.

These results strongly underscore the superiority of dynamic decision-making mechanisms, such as those used in DrugMCTS, over traditional static prediction methods like deep learning models. They also highlight the limitations of general-purpose large models in zero-shot settings for drug repositioning tasks, especially when retrieval-augmented prompting introduces noise or irrelevant context. This further emphasizes the importance of structured reasoning, domain-specific knowledge integration, and information filtering in such applications.

Furthermore, to demonstrate the interpretability and transparency of our approach, we provide a detailed case study in the Appendix Section~\ref{sec:case}. It illustrates how the model predicts an interaction between Equol and the CXC chemokine receptor 3, including the complete step-by-step reasoning process during a specific MCTS rollout.

\begin{table*}[!h]
\caption{Performance comparison on DrugBank and KIBA datasets}
\centering
\begin{adjustbox}{width=\linewidth}
\begin{tabular}{lcccc}
\toprule
\textbf{Model} & \textbf{Size} & \textbf{Dynamic Update} & \textbf{DrugBank} & \textbf{KIBA} \\
\midrule
\multicolumn{5}{l}{\textit{General Models (GM)}} \\
GPT-4o-mini & $\sim$8B & \texttimes & 0.1552 & 0.2580 \\
Deepseek-R1 & 37B\textsuperscript{a} & \texttimes & 0.1619 & 0.2645 \\
\midrule
\addlinespace
\multicolumn{5}{l}{\textit{GM + RAG}} \\
GPT-4o-mini & $\sim$8B & \checkmark & 0.1519 & 0.2252 \\
Deepseek-R1 & 37B\textsuperscript{a} & \checkmark & 0.1259 & 0.2173 \\
\midrule
\addlinespace
\multicolumn{5}{l}{\textit{DL Models}} \\
DL models & 8M-12M & \texttimes & 0.2364 & 0.3216 \\
\midrule
\addlinespace
\multicolumn{5}{l}{\textbf{DrugMCTS (Ours)}} \\
Selection = GT count & 7B & \checkmark & 0.4466 & 0.4224 \\
Selection = GT + 3 & 7B & \checkmark & 0.5534 & 0.4924 \\
\bottomrule
\end{tabular}
\end{adjustbox}

\vspace{4pt}
\footnotesize{
\textsuperscript{a}Activation-aware model size. 
GT = Ground Truth count. 
Dynamic update: \checkmark~Yes, \texttimes~No.
}
\label{tab:exp}
\end{table*}

\section{Ablation Studies}
\subsection{Settings}
In our ablation studies, we aim to investigate several key aspects:
\begin{itemize}
\item Whether the MCTS algorithm can improve model accuracy.
\item The effectiveness of our proposed data processing pipelines for scientific data, a hybrid of scientific and general data, and general data.
\item The efficacy of the combined relative and absolute reward calculation method.
\end{itemize}
To address these questions, we conducted the following experiments:

\begin{itemize}
\item \textbf{$S_1$ Baseline Setup.} Provide only the query molecule, all proteins in the protein pool, and the types of their pocket. Do not use the MCTS algorithm.
\item \textbf{$S_2$ Enhanced Information (EI) Setup.} On top of the baseline setup, add the detailed pockets information and literature information for all proteins and structural and chemical properties of the query molecule.
\item \textbf{$S_3$ Molecule Analysis Exclusion (MAE).} Conduct the MCTS process while excluding the molecule analysis action.
\item \textbf{$S_4$ Interaction Analysis Exclusion (IAE).} Conduct the MCTS process while excluding the interaction analysis action.
\item \textbf{$S_5$ Dual Exclusion (DE).} Conduct the MCTS process while excluding both the molecule analysis and interaction analysis actions.
\item \textbf{$S_6$ Relative Reward (RR) Only.} During the MCTS process, compute only the relative rewards without considering the absolute rewards.
\end{itemize}

Since excluding certain actions in MCTS can occasionally reduce the number of generated nodes, sometimes below the number of ground-truth targets, we ensure fair comparison by setting the MCTS-based baselines to generate four candidate nodes during the Protein Selection Action, instead of a single node used in the full pipeline.

\begin{table*}[h]
\centering
\caption{Performance comparison on DrugBank and KIBA datasets using Qwen7b (Top-k/Top-k+3 accuracy).}
\label{tab:ablation_setups}
\setlength{\tabcolsep}{4pt} 
\small 
\begin{tabular}{@{}lccccccc@{}} 
\toprule
\textbf{Setup} & \textbf{$S_1$ Baseline} & \textbf{$S_2$ EI} & \textbf{$S_3$ MAE} & \textbf{$S_4$ IAE} & \textbf{$S_5$ DE} & \textbf{$S_6$ RR} & \textbf{DrugMCTS} \\
\midrule
DrugBank & 0.1285 & 0.1586 & 0.3879/0.4677 & 0.3946/0.5119 & 0.3472/0.3617 & 0.4320/0.5527 & \textbf{0.4466/0.5534} \\
KIBA     & 0.2284 & 0.2452 & 0.3772/0.4352 & 0.3846/0.4491 & 0.3189/0.3264 & 0.4193/0.4861 & \textbf{0.4224/0.4924} \\
\bottomrule
\end{tabular}
\end{table*}

\subsection{Ablation Studies Results}
The ablation study results (Table~\ref{tab:ablation_setups}) clearly demonstrate the effectiveness of our proposed framework components. First, comparing $S_1$ (Baseline Setup) with $S_2$ (Enhanced Information Setup) shows that providing richer contextual information, including detailed pocket features, literature descriptions, structural and chemical properties, does improve performance to some extent (e.g., from 12.85\% to 15.86\% on DrugBank). However, the most significant improvement is observed when the MCTS algorithm is introduced in combination with these enhancements. Experimental settings that incorporate MCTS ($S_3$–$S_6$ and Final Result) consistently achieve much higher performance than $S_1$ or $S_2$, indicating that while richer input representations are beneficial, it is the MCTS-based reasoning process that plays the central role in boosting model accuracy.

Second, by analyzing $S_3$ (Molecule Analysis Exclusion), $S_4$ (Interaction Analysis Exclusion), and $S_5$ (Dual Exclusion), we observe a consistent drop in performance when either or both of the analysis modules are removed. For instance, on the DrugBank dataset, removing molecule analysis alone leads to a decrease from 44.66\% (Final Result) to 38.79\%, while removing interaction analysis results in a drop to 39.46\%. The dual exclusion further reduces performance to 34.72\%, demonstrating that each data processing step contributes meaningfully to the overall effectiveness of the system. This supports our hypothesis that the proposed hybrid data processing pipeline, incorporating both molecular and interaction-level analyses, is essential for capturing comprehensive contextual information.

Third, regarding the reward mechanism, the comparison between $S_6$ (Relative Reward Only) and the full reward setting (Final Result) shows that the combined use of relative and absolute rewards does not lead to a significant improvement in performance. One possible explanation is that prior steps, including candidate protein and molecule selection, have already filtered out most irrelevant options, leaving a refined set of high-quality reference proteins. As a result, the likelihood of encountering scenarios where none of the candidates interact with the query molecule becomes rare, reducing the added value of the absolute reward component.

In summary, these findings confirm the importance of the MCTS algorithm, the multi-step data processing pipeline, and the overall design of the retrieval-augmented reasoning framework in achieving strong performance in molecular-target interaction prediction.

\subsection{Computation Overhead Results}
For inference time scaling, the discussion typically revolves around two key aspects: the trade-off between model performance improvement and additional computational overhead. Therefore, in this section, we first analyze the impact of different rollout numbers on model performance and then compare the performance-overhead profile with baseline models.

Our analysis (Figure~\ref{fig:rollout} and Figure~\ref{fig:trade-off}) reveals that when the number of rollouts increases from 8 to 12, both Top-K and Top-K+3 metrics exhibit significant improvements across the two datasets. However, further increasing the rollout count from 12 to 24 only yields a notable gain in the Top-K+3 metric on the KIBA dataset, while other scenarios show either marginal or even negative improvements. Consequently, to balance computational cost and model performance, we ultimately adopt rollout=12 for our experiments. Compared to baseline models, our approach not only achieves the highest recall scores but also demonstrates superior cost efficiency, as evidenced by its position on the Pareto front.

\section{Conclusion}
Our DrugMCTS framework revolutionizes drug repositioning by integrating multi-agent collaboration (five specialized agents), hybrid data processing (scientific → hybrid → general), and MCTS to enable the lightweight Qwen2.5-7B-Instruct model to outperform Deepseek-R1 by more than 20\% recall on DrugBank/KIBA datasets. The system achieves 55.34\% recall via dynamic Top-K+3 selection, validated by 1,221 experimental interactions and case studies like Equol-CXCR3 binding (docking score: -8.4 kcal/mol). This work establishes a template for LLM-powered scientific discovery beyond drug-target prediction.

\section{Limitations}
While DrugMCTS demonstrates significant improvements over baseline models, several limitations highlight opportunities for further optimization:

\begin{itemize}
    \item Despite achieving more than 20\% recall gains over Deepseek-R1 (Table~\ref{tab:exp}), the absolute performance (55.34\% recall) suggests untapped optimization potential. The plateau in gains beyond 12 rollouts indicates diminishing returns from current MCTS configurations. 
    \item Current predictions primarily leverage PDB-derived binding pocket data (Section 2.4), omitting higher-order biological context. Future work may augment the framework with knowledge graph or Pathway activation score.
    \item The combined relative/absolute reward system (Eq. 6) yields only around 1\% improvement over relative-only rewards, suggesting the necessities of a more effective reward system. 
\end{itemize}

\clearpage          
\onecolumn  
\appendix 
\section{Appendix}
{\footnotesize
\begin{algorithm}[!ht]
\caption{A single MCTS rollout within the DrugMCTS framework.}
\label{alg:drugmcts}
\begin{algorithmic}[1]
\Require Query molecule $M_{qm}$
\Ensure Selected protein $P_s$, final reward $R_{\text{final}}(p^*)$

\State \textbf{// A\textsubscript{1} Retrieval Action}
\State Compute Tanimoto and ChemBERTa cosine similarities
\State $M_{cm} \gets \text{dedup}(\text{Top}_{10}(\text{sim}_T) \cup \text{Top}_{10}(\text{sim}_C))$
\State $P_{cp} \gets$ Proteins interacting with $M_{cm}$

\State \textbf{// A\textsubscript{2} Molecule Analysis Action}
\State $(C_{q,s}, C_{q,phy}) \gets \text{RDKit+PubChem}(M_{qm})$
\State $R_{qm} \gets$ Generate textual summary from $(C_{q,s}, C_{q,phy})$

\State \textbf{// A\textsubscript{3} Molecule Selection Action}
\State For all $m \in M_{cm}$: extract $(C_{c,s}, C_{c,phy})$ using MA agent
\State $M_{rm} \gets \text{MS Agent}(M_{qm}, R_{qm}, M_{cm}, C_{c,s}, C_{c,phy})$
\State $P_{rp} \gets$ Proteins in $P_{cp}$ interacting with $M_{rm}$

\State \textbf{// A\textsubscript{4} Interaction Analysis Action}
\State $D_{bp} \gets$ Extract pockets from PDBs via PLIP
\State $L_{rp} \gets$ Retrieve protein literature from PubMed
\State $R_{ia} \gets \text{IA Agent}(M_{rm}, P_{rp}, D_{bp}, L_{rp})$

\State \textbf{// A\textsubscript{5} Protein Selection Action}
\State $P_s \gets \text{Decision Agent}(M_{qm}, R_{qm}, M_{rm}, P_{rp}, D_{bp}, R_{ia})$

\State \textbf{// A\textsubscript{6} End Action}
\State Perform backpropagation in MCTS from end node

\State \textbf{// Reward Calculation}
\State Query model $k$ times to get selections $\{p^{(1)}, \dots, p^{(k)}\}$
\State $p^* \gets$ most frequently selected candidate
\State $R_{\text{relative}}(p^*) \gets \frac{\#\text{times } p^* \text{ is selected}}{k}$

\State Feed $(P_s, D_{bp}, L_{rp}, M_{qm}, R_{qm})$ to decision model $k$ times
\State $R_{\text{absolute}}(p^*) \gets \frac{\#\text{‘yes’ responses}}{k}$

\State \textbf{Return:} $P_s$, $R_{\text{final}}(p^*) = \frac{R_{\text{relative}}(p^*) + R_{\text{absolute}}(p^*)}{2}$

\end{algorithmic}
\end{algorithm}
}

\begin{algorithm}[!htbp]
\caption{MCTS for DrugMCTS}
\label{alg:mcts}
\begin{algorithmic}[1]
\Require Query molecule $M_{qm}$, number of simulations $N$
\Ensure Selected protein $P_s$

\State Initialize root node $s_0$ with $M_{qm}$
\For{$i = 1$ to $N$}
    \State $s_{\text{leaf}} \gets$ \textbf{Select}$(s_0)$ using UCT
    \State $s_{\text{child}} \gets$ \textbf{Expand}$(s_{\text{leaf}})$
    \State $R_i \gets$ \textbf{Rollout}$(s_{\text{child}})$ \Comment{See Algorithm~\ref{alg:drugmcts}}
    \State \textbf{Backpropagate} reward $R_i$ to update tree
\EndFor
\State \Return $P_s \gets \textbf{BestChild}(s_0)$ based on average reward
\end{algorithmic}
\end{algorithm}

\subsection{Dataset Construction}
\label{sec:dataset}
\textbf{Experimental Dataset.} We first extracted all molecules from the original dataset and computed both Tanimoto similarity and cosine similarity between each pair. For each molecule, we selected the top 10 most similar molecules based on each metric, merged the results, and removed duplicates to form the candidate molecule set $M_{cm}$. Each unique query molecule paired with its corresponding $M_{cm}$ constitutes one problem instance. These problem instances were further filtered according to the following criteria:
\begin{itemize}
\item For each query molecule, the number of associated interacting proteins must be between 2 and 10.
\item For each candidate molecule in $M_{cm}$, the number of associated interacting proteins must be between 2 and 4.
\item The total number of candidate molecules per query must not exceed 15.
\end{itemize}

\begin{table}[!htbp]
\centering
\caption{Experimental Dataset Statistics}
\label{tab:dataset_stats}
\begin{tabular}{l *{5}{S[table-format=4.0]}} 
\toprule
{Dataset} & {Processed Points} & {All Proteins} & {Ground Truth} & {Unique Molecules} & {All Molecules} \\
\midrule
DrugBank       & 788  & 22508 & 1595 & 1304 & 7717 \\
KIBA     & 626  & 23849 & 1664 &  752 & 6219 \\
\bottomrule
\end{tabular}
\end{table}

We then extracted all proteins that interact with any molecule in $M_{cm}$, denoted as $P_{cp}$ (as mentioned in Section 2.4). Additionally, we collected all proteins that directly interact with the query molecule. The intersection of these two protein sets was defined as the ground truth set.
The ground truth set must satisfy the following constraints:

\begin{itemize}
\item Its size must be between 1 and 5.
\item Its size must not exceed 70\% of the size of the candidate protein set.
\end{itemize}

\textbf{Baseline Dataset.} This dataset is derived from the Experimental Dataset, with the following modifications:

\begin{itemize}
\item Only the query molecules, the candidate protein set $P_{cp}$, and the ground truth set are retained.
\item All candidate molecules $M_{cm}$ are removed.
\end{itemize}

\begin{table}[!htbp]
\centering
\caption{Baseline Dataset Statistics}
\label{tab:baseline_stats}
\scriptsize  
\setlength{\tabcolsep}{20pt}  
\begin{tabular}{l S[table-format=4.0] S[table-format=5.0] S[table-format=2.2]}
\toprule
{Dataset} & {Ground Truth} & {All Proteins} & {Ratio (\%)} \\
\midrule
DB       & 1595  & 14654 & 10.88 \\
KIBA     & 1664  & 10593 & 15.71 \\
\bottomrule
\end{tabular}
\end{table}

\textbf{Deep Learning Dataset.} From the original dataset, we extracted all data instances involving the query molecules from the Experimental Dataset to form the test set. The remaining data were used as the training set. 

\begin{table}[!htbp]
\centering
\caption{Deep Learning Dataset Statistics}
\resizebox{0.8\textwidth}{!}{
\begin{tabular}{l l r r r}  
\toprule
\textbf{Dataset} & \textbf{Split} & \textbf{Negative} & \textbf{Positive} & \textbf{Total} \\
\midrule
\multirow{2}{*}{DrugBank} 
 & Train & 14,787 & 12,428 & 27,215 \\
 & Test  &  1,960 &  4,111 &  6,071 \\
\midrule
\multirow{2}{*}{KIBA} 
 & Train & 62,553 & 17,350 & 79,903 \\
 & Test  & 31,643 &  4,804 & 36,447 \\
\bottomrule
\end{tabular}
}
\label{tab:dl_dataset}
\end{table}

\begin{figure}[!htbp]
  \centering
  \begin{subfigure}[b]{0.9\textwidth} 
    \includegraphics[width=\linewidth]{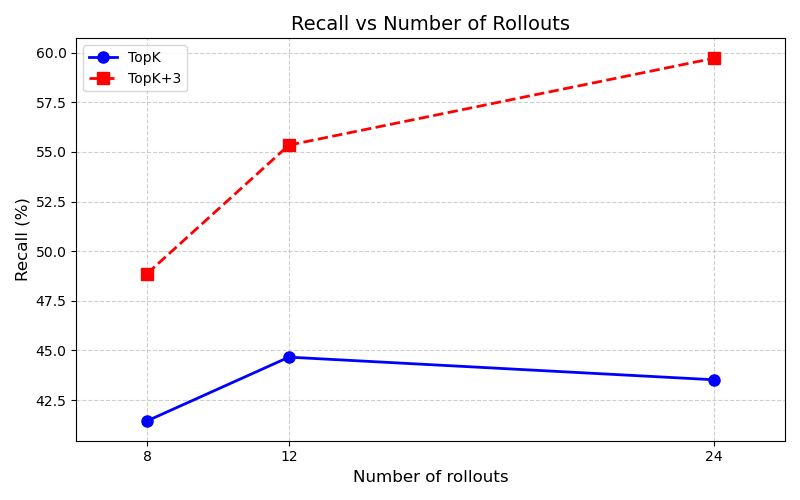}
    \caption{}
    \label{fig:DB_r}
  \end{subfigure}
  
  \vspace{10pt} 
  
  \begin{subfigure}[b]{0.9\textwidth}
    \includegraphics[width=\linewidth]{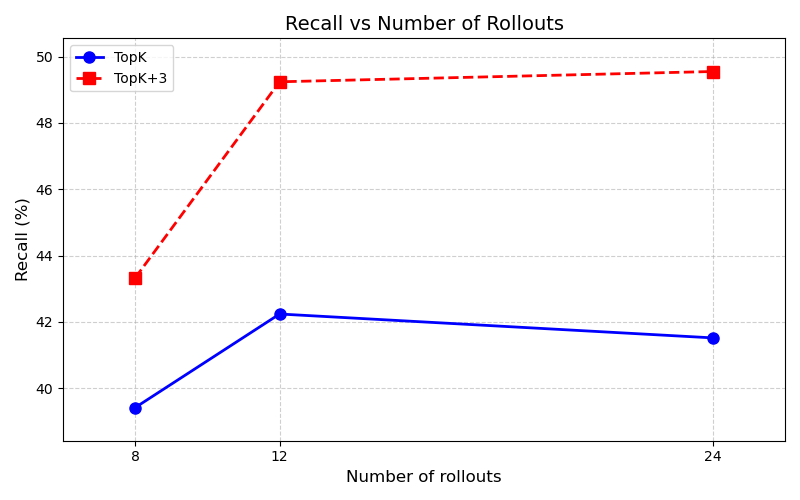}
    \caption{}
    \label{fig:KIBA_r}
  \end{subfigure}
  
  \caption{Number of rollouts vs Recall score on (a) DrugBank and (b) KIBA dataset}
  \label{fig:rollout}
\end{figure}

\begin{figure}[!htbp]
  \centering
  \begin{subfigure}[b]{0.9\textwidth} 
    \includegraphics[width=\linewidth]{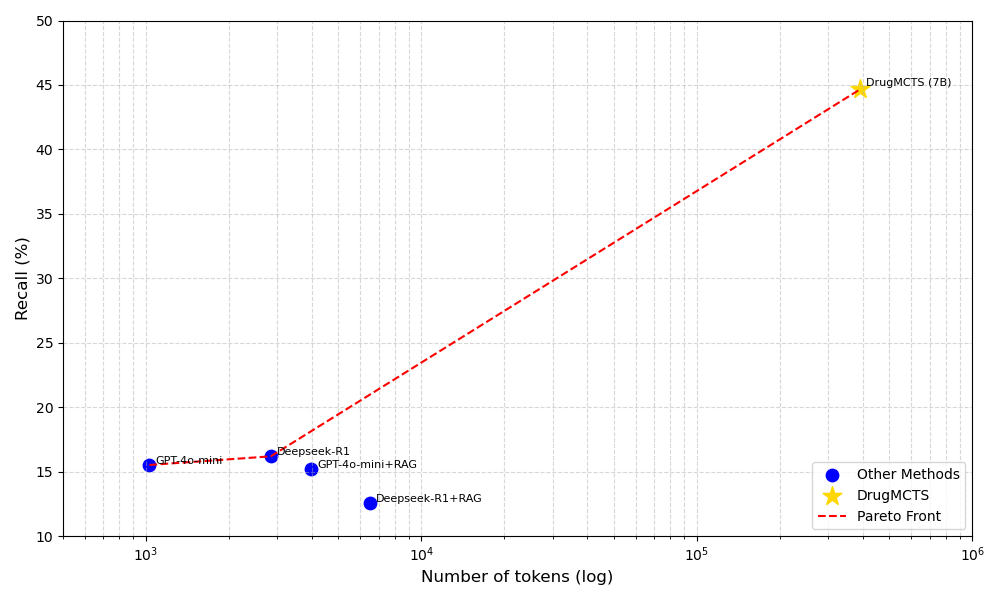}
    \caption{}
    \label{fig:DB_t}
  \end{subfigure}
  
  \vspace{10pt} 
  
  \begin{subfigure}[b]{0.9\textwidth}
    \includegraphics[width=\linewidth]{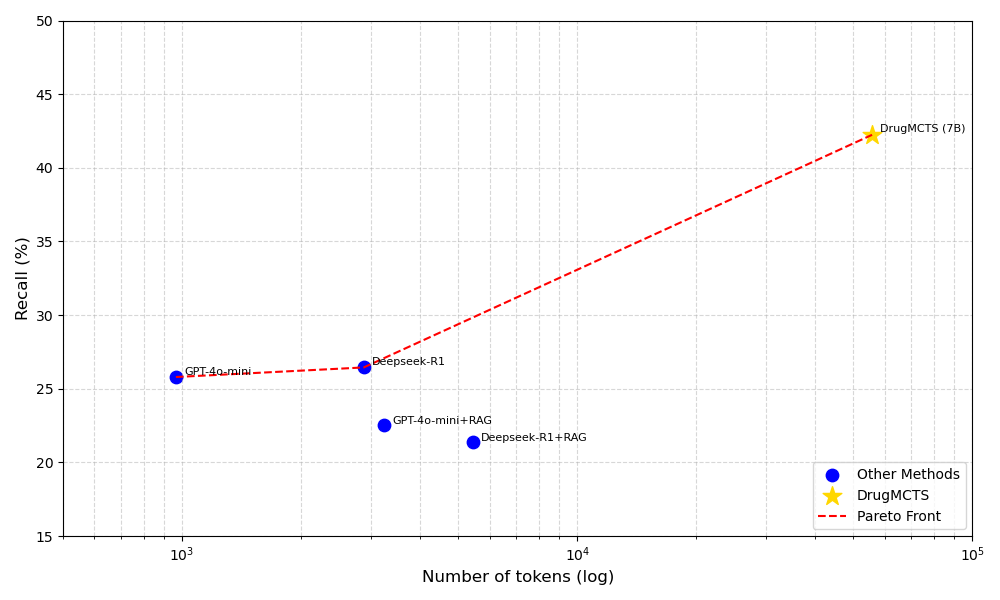}
    \caption{}
    \label{fig:KIBA_t}
  \end{subfigure}
  
  \caption{Number of tokens vs Recall score on (a) DrugBank and (b) KIBA dataset}
  \label{fig:trade-off}
\end{figure}

\subsection{Case Study}
\label{sec:case}
We manually selected a molecular-protein interaction with the highest self-consistency score that has never been previously reported: Equol (DrugBank ID: DB11674) and CXC chemokine receptor 3 (CXCR3, PDB ID: 8K2W). The binding affinity predicted by AutoDock Vina was -8.4 kcal/mol, indicating a strong potential interaction between the two. Visualization using PyMOL~\cite{yuan2017using} revealed that Equol can bind within one of the binding pockets of CXCR3 and form hydrogen bonds, as evidenced by the continuous red dots in the lower-right corner of Appendix Figure~\ref{fig:combined}(b). This observation is consistent with the reasoning generated by the large language model during the molecule analysis, protein selection, and absolute reward calculation stages, thereby validating the effectiveness of our framework.

\begin{figure}[!htbp]
  \centering
  \begin{subfigure}[b]{0.48\textwidth} 
    \includegraphics[width=\linewidth]{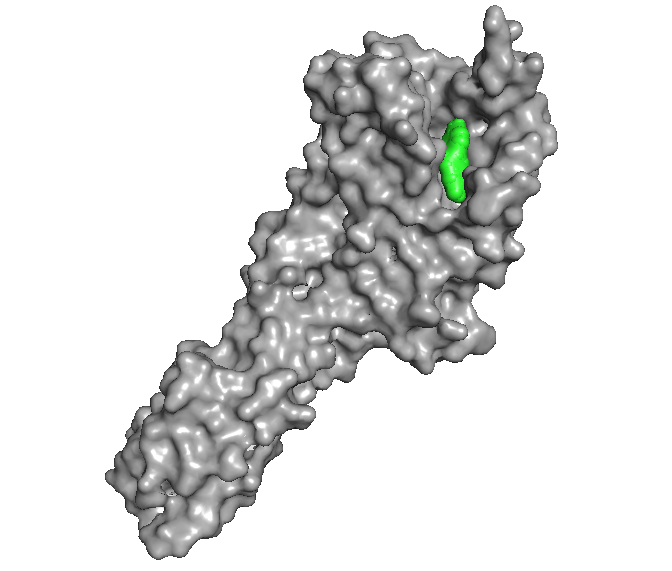}
    \caption{}
    \label{fig:global}
  \end{subfigure}
  \hfill 
  \begin{subfigure}[b]{0.48\textwidth}
    \includegraphics[width=\linewidth]{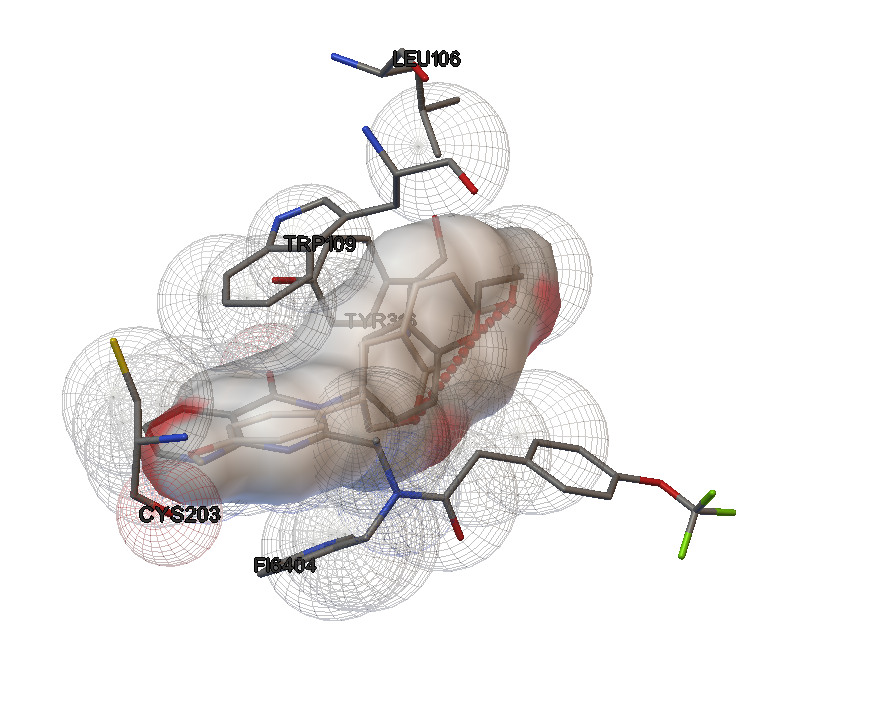}
    \caption{}
    \label{fig:local}
  \end{subfigure}
  \caption{Protein-ligand docking results between (a) global binding site overview and (b) detailed interaction view, computed by AutoDock Vina with default scoring function and visualized using PyMOL. }
  \label{fig:combined}
\end{figure}

\begin{figure}[!htbp]
  \centering
  \includegraphics[width=\textwidth]{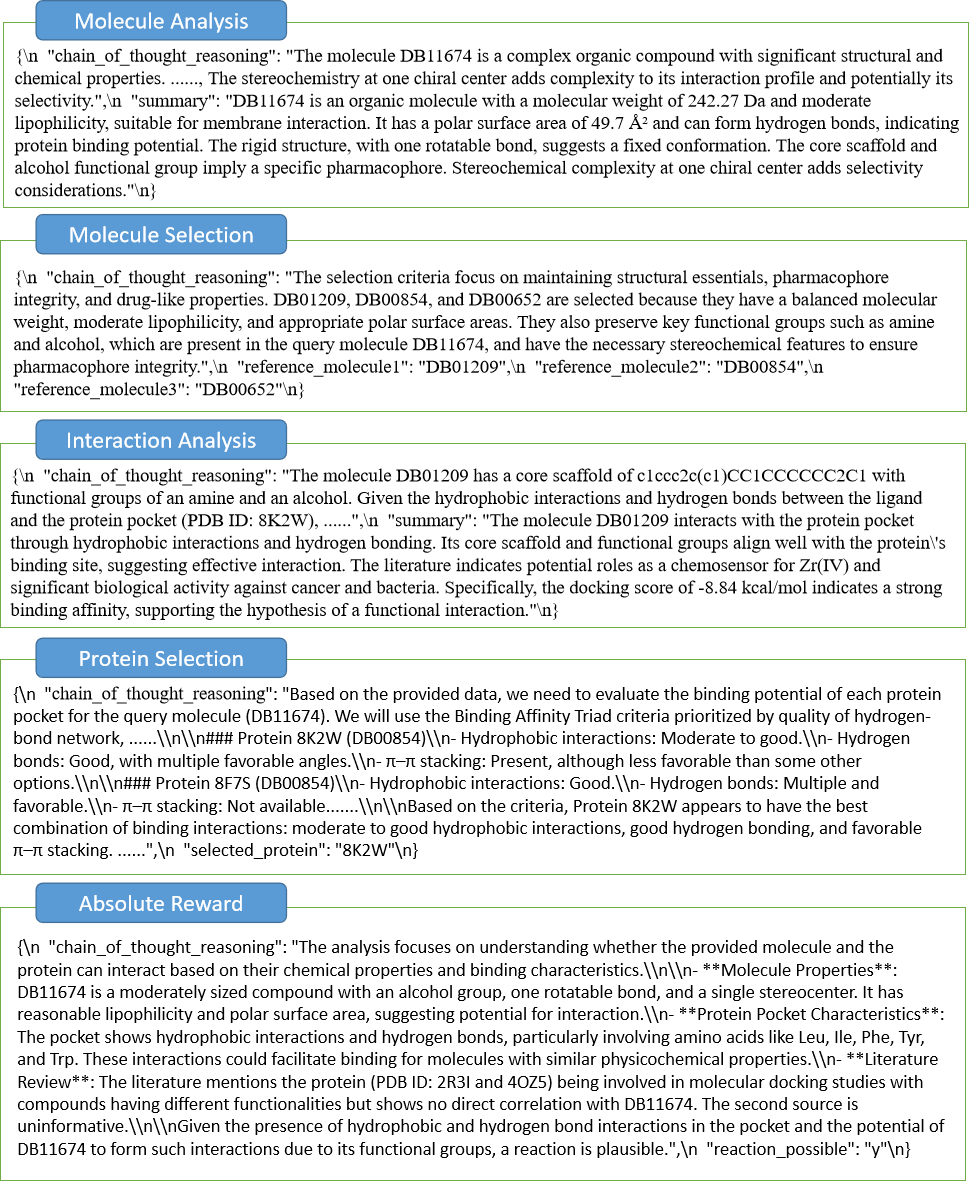}  
  \caption{Answers generated by the model}
  \label{fig:example}
\end{figure}


\begin{thebibliography}{10}

\bibitem{bajusz2015tanimoto}
D{\'a}vid Bajusz, Anita R{\'a}cz, and K{\'a}roly H{\'e}berger.
\newblock Why is tanimoto index an appropriate choice for fingerprint-based similarity calculations?
\newblock {\em Journal of cheminformatics}, 7:1--13, 2015.

\bibitem{chaslot2010monte}
Guillaume Maurice Jean-Bernard~Chaslot Chaslot.
\newblock Monte-carlo tree search.
\newblock 2010.

\bibitem{che2025csstep}
Xinhao Che, Yujing Zhao, Qilei Liu, Fang Yu, Hanyu Gao, and Lei Zhang.
\newblock Csstep: Step-by-step exploration of the chemical space of drug molecules via multi-agent and multi-stage reinforcement learning.
\newblock {\em Chemical Engineering Science}, page 122048, 2025.

\bibitem{chen2025improving}
Yiqun Chen, Lingyong Yan, Weiwei Sun, Xinyu Ma, Yi~Zhang, Shuaiqiang Wang, Dawei Yin, Yiming Yang, and Jiaxin Mao.
\newblock Improving retrieval-augmented generation through multi-agent reinforcement learning.
\newblock {\em arXiv preprint arXiv:2501.15228}, 2025.

\bibitem{couetoux2011continuous}
Adrien Cou{\"e}toux, Jean-Baptiste Hoock, Nataliya Sokolovska, Olivier Teytaud, and Nicolas Bonnard.
\newblock Continuous upper confidence trees.
\newblock In {\em Learning and Intelligent Optimization: 5th International Conference, LION 5, Rome, Italy, January 17-21, 2011. Selected Papers 5}, pages 433--445. Springer, 2011.

\bibitem{edwards2024molcap}
Carl Edwards, Ziqing Lu, Ehsan Hajiramezanali, Tommaso Biancalani, Heng Ji, and Gabriele Scalia.
\newblock Molcap-arena: A comprehensive captioning benchmark on language-enhanced molecular property prediction.
\newblock {\em arXiv preprint arXiv:2411.00737}, 2024.

\bibitem{guo2025deepseek}
Daya Guo, Dejian Yang, Haowei Zhang, Junxiao Song, Ruoyu Zhang, Runxin Xu, Qihao Zhu, Shirong Ma, Peiyi Wang, Xiao Bi, et~al.
\newblock Deepseek-r1: Incentivizing reasoning capability in llms via reinforcement learning.
\newblock {\em arXiv preprint arXiv:2501.12948}, 2025.

\bibitem{huey2012using}
Ruth Huey, Garrett~M Morris, Stefano Forli, et~al.
\newblock Using autodock 4 and autodock vina with autodocktools: a tutorial.
\newblock {\em The Scripps Research Institute Molecular Graphics Laboratory}, 10550(92037):1000, 2012.

\bibitem{hurst2024gpt}
Aaron Hurst, Adam Lerer, Adam~P Goucher, Adam Perelman, Aditya Ramesh, Aidan Clark, AJ~Ostrow, Akila Welihinda, Alan Hayes, Alec Radford, et~al.
\newblock Gpt-4o system card.
\newblock {\em arXiv preprint arXiv:2410.21276}, 2024.

\bibitem{hutter2025lost}
Jan Hutter, David Rau, Maarten Marx, and Jaap Kamps.
\newblock Lost but not only in the middle: Positional bias in retrieval augmented generation.
\newblock In {\em European Conference on Information Retrieval}, pages 247--261. Springer, 2025.

\bibitem{inoue2025drugagent}
Yoshitaka Inoue, Tianci Song, Xinling Wang, Augustin Luna, and Tianfan Fu.
\newblock Drugagent: Multi-agent large language model-based reasoning for drug-target interaction prediction.
\newblock In {\em ICLR 2025 Workshop on Machine Learning for Genomics Explorations}, 2025.

\bibitem{knox2024drugbank}
Craig Knox, Mike Wilson, Christen~M Klinger, Mark Franklin, Eponine Oler, Alex Wilson, Allison Pon, Jordan Cox, Na~Eun Chin, Seth~A Strawbridge, et~al.
\newblock Drugbank 6.0: the drugbank knowledgebase for 2024.
\newblock {\em Nucleic acids research}, 52(D1):D1265--D1275, 2024.

\bibitem{lee2019deepconv}
Ingoo Lee, Jongsoo Keum, and Hojung Nam.
\newblock Deepconv-dti: Prediction of drug-target interactions via deep learning with convolution on protein sequences.
\newblock {\em PLoS computational biology}, 15(6):e1007129, 2019.

\bibitem{lee2025rag}
Namkyeong Lee, Edward De~Brouwer, Ehsan Hajiramezanali, Tommaso Biancalani, Chanyoung Park, and Gabriele Scalia.
\newblock Rag-enhanced collaborative llm agents for drug discovery.
\newblock {\em arXiv preprint arXiv:2502.17506}, 2025.

\bibitem{li2025alpha}
Boyan Li, Jiayi Zhang, Ju~Fan, Yanwei Xu, Chong Chen, Nan Tang, and Yuyu Luo.
\newblock Alpha-sql: Zero-shot text-to-sql using monte carlo tree search.
\newblock {\em arXiv preprint arXiv:2502.17248}, 2025.

\bibitem{liu2023lost}
Nelson~F Liu, Kevin Lin, John Hewitt, Ashwin Paranjape, Michele Bevilacqua, Fabio Petroni, and Percy Liang.
\newblock Lost in the middle: How language models use long contexts.
\newblock {\em arXiv preprint arXiv:2307.03172}, 2023.

\bibitem{liu2024drugagent}
Sizhe Liu, Yizhou Lu, Siyu Chen, Xiyang Hu, Jieyu Zhao, Yingzhou Lu, and Yue Zhao.
\newblock Drugagent: Automating ai-aided drug discovery programming through llm multi-agent collaboration.
\newblock {\em arXiv preprint arXiv:2411.15692}, 2024.

\bibitem{luo2025does}
Kun Luo, Zheng Liu, Peitian Zhang, Hongjin Qian, Jun Zhao, and Kang Liu.
\newblock Does rag really perform bad for long-context processing?
\newblock {\em arXiv preprint arXiv:2502.11444}, 2025.

\bibitem{nguyen2019toward}
Cuong~V Nguyen, Alessandro Achille, Michael Lam, Tal Hassner, Vijay Mahadevan, and Stefano Soatto.
\newblock Toward understanding catastrophic forgetting in continual learning.
\newblock {\em arXiv preprint arXiv:1908.01091}, 2019.

\bibitem{nguyen2023perceiver}
Ngoc-Quang Nguyen, Gwanghoon Jang, Hajung Kim, and Jaewoo Kang.
\newblock Perceiver cpi: a nested cross-attention network for compound--protein interaction prediction.
\newblock {\em Bioinformatics}, 39(1):btac731, 2023.

\bibitem{nguyen2021graphdta}
Thin Nguyen, Hang Le, Thomas~P Quinn, Tri Nguyen, Thuc~Duy Le, and Svetha Venkatesh.
\newblock Graphdta: predicting drug--target binding affinity with graph neural networks.
\newblock {\em Bioinformatics}, 37(8):1140--1147, 2021.

\bibitem{csakar2025maximizing}
Tolga {\c{S}}akar and Hakan Emekci.
\newblock Maximizing rag efficiency: A comparative analysis of rag methods.
\newblock {\em Natural Language Processing}, 31(1):1--25, 2025.

\bibitem{song2025llm}
Kevin Song, Andrew Trotter, and Jake~Y Chen.
\newblock Llm agent swarm for hypothesis-driven drug discovery.
\newblock {\em arXiv preprint arXiv:2504.17967}, 2025.

\bibitem{swain2014pubchempy}
Matt Swain.
\newblock Pubchempy documentation.
\newblock In {\em PubChemPy documentation}. 2014.

\bibitem{tang2014making}
Jing Tang, Agnieszka Szwajda, Sushil Shakyawar, Tao Xu, Petteri Hintsanen, Krister Wennerberg, and Tero Aittokallio.
\newblock Making sense of large-scale kinase inhibitor bioactivity data sets: a comparative and integrative analysis.
\newblock {\em Journal of chemical information and modeling}, 54(3):735--743, 2014.

\bibitem{team2024qwen2}
Qwen Team.
\newblock Qwen2 technical report.
\newblock {\em arXiv preprint arXiv:2412.15115}, 2024.

\bibitem{van2025assessment}
Joren Van~Herck, Mar{\'\i}a~Victoria Gil, Kevin~Maik Jablonka, Alex Abrudan, Andy~S Anker, Mehrdad Asgari, Ben Blaiszik, Antonio Buffo, Leander Choudhury, Clemence Corminboeuf, et~al.
\newblock Assessment of fine-tuned large language models for real-world chemistry and material science applications.
\newblock {\em Chemical science}, 16(2):670--684, 2025.

\bibitem{wang2022self}
Xuezhi Wang, Jason Wei, Dale Schuurmans, Quoc Le, Ed~Chi, Sharan Narang, Aakanksha Chowdhery, and Denny Zhou.
\newblock Self-consistency improves chain of thought reasoning in language models.
\newblock {\em arXiv preprint arXiv:2203.11171}, 2022.

\bibitem{wei2024drugrealign}
Jinhang Wei, Linlin Zhuo, Xiangzheng Fu, XiangXiang Zeng, Li~Wang, Quan Zou, and Dongsheng Cao.
\newblock Drugrealign: a multisource prompt framework for drug repurposing based on large language models.
\newblock {\em BMC biology}, 22(1):226, 2024.

\bibitem{white2020pubmed}
Jacob White.
\newblock Pubmed 2.0.
\newblock {\em Medical reference services quarterly}, 39(4):382--387, 2020.

\bibitem{yang2025mhmg}
Zerui Yang, Yinqiao Li, Yudai Matsuda, and Linqi Song.
\newblock mhmg-dti: A drug-target interaction prediction framework combining modified hierarchical molecular graphs and improved convolutional block attention module.
\newblock In {\em Trends and Applications in Knowledge Discovery and Data Mining: PAKDD 2025 Workshops, ADUR, FairPC, GLFM, PM4B and RAFDA, Sydney, NSW, Australia, June 10--13, 2025, Proceedings}, volume 15835, page 191. Springer Nature, 2025.

\bibitem{yang2024iresnetdm}
Zerui Yang, Wei Shao, Yudai Matsuda, and Linqi Song.
\newblock iresnetdm: An interpretable deep learning approach for four types of dna methylation modification prediction.
\newblock {\em Computational and Structural Biotechnology Journal}, 23:4214--4221, 2024.

\bibitem{ye2025drugassist}
Geyan Ye, Xibao Cai, Houtim Lai, Xing Wang, Junhong Huang, Longyue Wang, Wei Liu, and Xiangxiang Zeng.
\newblock Drugassist: A large language model for molecule optimization.
\newblock {\em Briefings in Bioinformatics}, 26(1):bbae693, 2025.

\bibitem{yuan2017using}
Shuguang Yuan, HC~Stephen Chan, and Zhenquan Hu.
\newblock Using pymol as a platform for computational drug design.
\newblock {\em Wiley Interdisciplinary Reviews: Computational Molecular Science}, 7(2):e1298, 2017.

\bibitem{zhang2025rag2mol}
Peidong Zhang, Xingang Peng, Rong Han, Ting Chen, and Jianzhu Ma.
\newblock Rag2mol: Structure-based drug design based on retrieval augmented generation.
\newblock {\em Briefings in Bioinformatics}, 26(3):bbaf265, 2025.

\bibitem{zhang2024fine}
Wei Zhang, Qinggong Wang, Xiangtai Kong, Jiacheng Xiong, Shengkun Ni, Duanhua Cao, Buying Niu, Mingan Chen, Yameng Li, Runze Zhang, et~al.
\newblock Fine-tuning large language models for chemical text mining.
\newblock {\em Chemical Science}, 15(27):10600--10611, 2024.

\bibitem{zhao2019attentiondta}
Qichang Zhao, Fen Xiao, Mengyun Yang, Yaohang Li, and Jianxin Wang.
\newblock Attentiondta: prediction of drug--target binding affinity using attention model.
\newblock In {\em 2019 IEEE international conference on bioinformatics and biomedicine (BIBM)}, pages 64--69. IEEE, 2019.

\bibitem{zheng2025large}
Yizhen Zheng, Huan~Yee Koh, Jiaxin Ju, Anh~TN Nguyen, Lauren~T May, Geoffrey~I Webb, and Shirui Pan.
\newblock Large language models for scientific discovery in molecular property prediction.
\newblock {\em Nature Machine Intelligence}, pages 1--11, 2025.

\bibitem{zheng2024large}
Yizhen Zheng, Huan~Yee Koh, Maddie Yang, Li~Li, Lauren~T May, Geoffrey~I Webb, Shirui Pan, and George Church.
\newblock Large language models in drug discovery and development: From disease mechanisms to clinical trials.
\newblock {\em arXiv preprint arXiv:2409.04481}, 2024.

\end{thebibliography}
\end{document}